\documentclass[runningheads]{llncs}
\usepackage{amsmath,amssymb,amsfonts}
\usepackage{algorithmic}
\usepackage{algorithm}

\usepackage{stix}
\usepackage{wasysym}

\usepackage{subfig}
\usepackage{booktabs}
\usepackage{graphicx}
\usepackage{textcomp}
\usepackage{xcolor}

\usepackage{color, colortbl}
\definecolor{LightCyan}{rgb}{0.88,1,1}

\title{\LARGE \bf
AdaGAT: Adaptive Guidance Adversarial Training for the Robustness of Deep Neural Networks
}

\begin{document}
%Enhancing Robustness via Pushing Guide Model Closer to Customized Adversarial Training
%Deep Neural Networks Robustness via Adaptive Guidance Adversarial Training
% \title{
% AdaGAT: Adaptive Guidance Adversarial Training for the Robustness of Deep Neural Networks
% %{\footnotesize \textsuperscript{*}Note: Sub-titles are not captured in Xplore and should not be used} \thanks{Identify applicable funding agency here. If none, delete this.}
% }

% \author{\IEEEauthorblockN{Zhenyu Liu}
% \IEEEauthorblockA{\textit{Newcastle University} \\
% Newcastle, UK \\
% z.liu48@newcastle.ac.uk}
% \and
% \IEEEauthorblockN{Huizhi Liang}
% \IEEEauthorblockA{\textit{Newcastle University} \\
% Newcastle, UK \\
% huizhi.liang@newcastle.ac.uk}
% \and
% \IEEEauthorblockN{Xinrun Li}
% \IEEEauthorblockA{\textit{Newcastle University} \\
% Newcastle, UK \\
% x.li169@newcastle.ac.uk}

% \linebreakand
% \IEEEauthorblockN{Rajiv Ranjan}
% \IEEEauthorblockA{\textit{Newcastle University} \\
% Newcastle, UK \\
% raj.ranjan@newcastle.ac.uk}
% \and
% \IEEEauthorblockN{Vaclav Snasel}
% \IEEEauthorblockA{\textit{Technical University of Ostrava} \\
% Ostrava, Czechia \\
% vaclav.snasel@vsb.cz}
%  \and
% \IEEEauthorblockN{Varun Ojha}
% \IEEEauthorblockA{\textit{Newcastle University} \\
% Newcastle, UK \\
% varun.ojha@newcastle.ac.uk}
% }
\author{
% %First Author\inst{1}\orcidID{0000-1111-2222-3333} \and
Zhenyu Liu\inst{1} \and
Huizhi Liang\inst{1} \and
Xinrun Li\inst{1} \and
Vaclav Snasel\inst{2}\and
Varun Ojha\inst{1}
% %Anonymous submission
}
% %
\authorrunning{Z. Liu et al.}

\institute{Newcastle University, Newcastle, UK \and
Technical University of Ostrava, Ostrava, Czech Republic
}
\maketitle

\begin{abstract}

Adversarial distillation (AD) is a knowledge distillation technique that facilitates the transfer of robustness from teacher deep neural network (DNN) models to lightweight target (student) DNN models, enabling the target models to perform better than only training the student model independently. Some previous works focus on using a small, learnable teacher (guide) model to improve the robustness of a student model. Since a learnable guide model starts learning from scratch, maintaining its optimal state for effective knowledge transfer during co-training is challenging. Therefore, we propose a novel \textit{Adaptive Guidance Adversarial Training} (AdaGAT) method. Our method, AdaGAT, dynamically adjusts the training state of the guide model to install robustness to the target model. Specifically, we develop two separate loss functions as part of the AdaGAT method, allowing the guide model to participate more actively in backpropagation to achieve its optimal state. We evaluated our approach via extensive experiments on three datasets: CIFAR-10, CIFAR-100, and TinyImageNet, using the WideResNet-34-10 model as the target model. Our observations reveal that appropriately adjusting the guide model within a certain accuracy range enhances the target model's robustness across various adversarial attacks compared to a variety of baseline models. Our code is available at the following link: {\color{teal}https://github.com/lusti-Yu/Adaptive-Gudiance-AT.git}.

\end{abstract}

% \begin{IEEEkeywords}
% Deep neural networks, knowledge distillation, adversarial attacks, adversarial defense, adversarial training, adversarial robustness
% \end{IEEEkeywords}

\section{Introduction}
Deep neural networks (DNNs) are critical in various real-world applications. From surveillance~\cite{yuan2024towards} and autonomous vehicles~\cite{zhang2022adversarial} to healthcare~\cite{ghosh2024clipsyntel}, DNNs have been proven most effective in solving real-world problems with high accuracy. However, studies have demonstrated that DNNs can be deceived by small perturbations, referred to as adversarial examples~\cite{szegedy2013intriguing}. Since this discovery, numerous techniques have been proposed to generate adversarial samples that can fool a DNN~\cite{goodfellow2014explaining,carlini2017towards}. Likewise, several adversarial defense techniques have been proposed to mitigate these vulnerabilities~\cite{rice2020overfitting,huang2023boosting}.

\begin{figure}[t]
\centering
\includegraphics[width=0.5\textwidth]{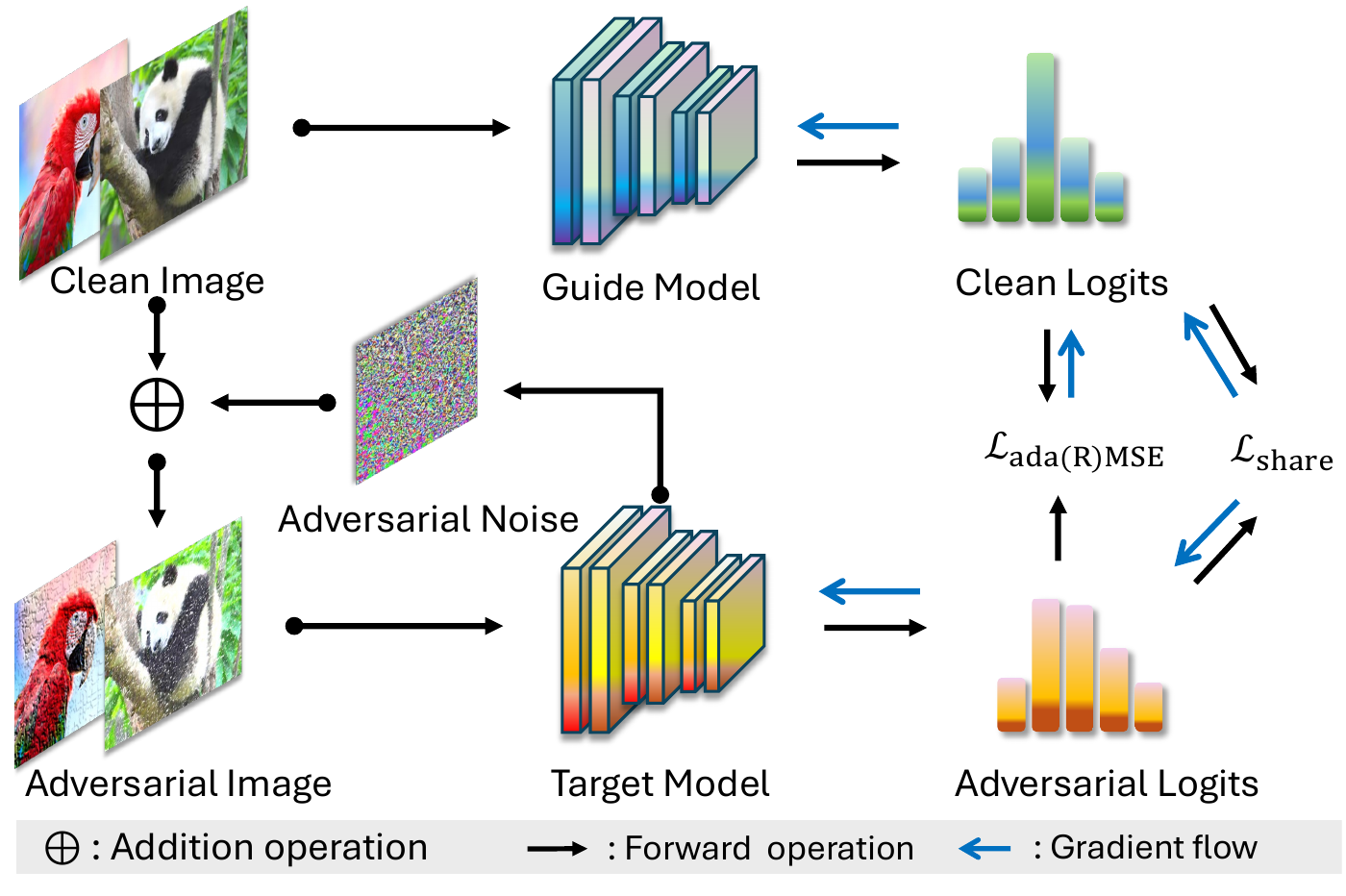}
\caption{Adaptive guidance adversarial training method (AdaGAT) framework introduces a comprehensive adversarial training approach by incorporating an additional strategic component: adaptive MSE loss. Specifically, $\mathcal{L}_{\text{CE}}$ is employed for clean classification training on the guide model using ground-truth labels. The proposed additional loss function is denoted as $\mathcal{L}_{\text{ada(R)MSE}}$, while $\mathcal{L}_{\text{share}}$ represents the adversarial training loss in LBGAT. These loss functions are combined to form the loss function of our AdaGAT method. Notably, in our proposed loss $\mathcal{L}_{\text{ada(R)MSE}}$, the target model's parameters are kept constant during backpropagation. Based on subsequent results and analysis,  our AdaGAT method is
shown to strengthen the robustness of the target model.}
\label{fig:MGLCM}
\end{figure}

Adversarial training has been among the most effective techniques to enhance the robustness of deep learning models~\cite{rice2020overfitting}. These techniques can be divided into three progressively advanced approaches. The \textit{first} category is single-model adversarial training, where adversarial techniques are applied directly to a single model. While this method lays the foundation for adversarial training, its ability to improve robustness is limited. The \textit{second} category of methods is adversarial distillation frameworks. These frameworks are employed for adversarial training to improve the robustness of a student model by utilizing pre-trained large DNN teacher models with superior robustness, i.e.,  the teacher DNN model transfers its knowledge to a smaller target DNN model~\cite{huang2023boosting,zi2021revisiting}. However, the robustness of the adversarial distillation approach is often limited to the transferability from the large teacher model to the smaller student model, and the training process of these robust pre-trained models generally lacks adaptability. Moving beyond these limitations, the \textit{third} category of methods explores using adaptive small models to assist larger target models in adversarial training~\cite{jia2022adversarial,cui2021learnable,liu2024dynamic,liu2025d2r}. This model adaptively adjusts according to the performance of the target model, offering a more flexible and effective approach to adversarial training.

Within the third category of adversarial training methods using the adaptive models, adversarial defense techniques, such as LBGAT~\cite{cui2021learnable}, employ a small learnable guide model to enhance the robustness of a large target model in adversarial settings. However, such collaborative training approaches face significant challenges, particularly in the challenge of adjusting the dynamic state of the guide model to maximize the target model's robustness. Our analysis in Section~\ref{Problem Definition And Motivation} reveals a strong correlation between the guide model's clean accuracy and the target model's robust performance during adversarial training. Based on this analysis, we argue that the clean accuracy of the guide model could potentially impact the robustness of the target model.

In this paper, we address this issue by adjusting the state of the guide model. Our approach aims to adjust the guide model’s state to achieve the best possible robustness performance of the target model. This ensures that the learnable guide model remains adaptive and closely synchronized with the target model’s state during adversarial training. Methodologically, we propose to regulate the guide model by incorporating an additional mean squared error (MSE) or root mean squared error (RMSE) loss with a stop-gradient operation (see Fig.~\ref{fig:MGLCM}). Specifically, MSE-like functions with a stop-gradient operation can adjust the guide model’s clean outputs to align more closely with the target model’s adversarial outputs, thereby enhancing the target model’s robustness. Moreover, RMSE plays a critical role in amplifying small output discrepancies, making the training process more sensitive to subtle differences, which improves adversarial robustness. The main contributions of this work can be summarized as follows:
\begin{itemize}

\item We identify the relationship between the target model's robust accuracy and the guide model's clean accuracy, demonstrating that a guide model with a certain degree of lower clean classification accuracy is positively correlated with the robustness of the target model. This observation allows us to uncover novel issues and propose innovative solutions.

\item We propose a novel \textit{Adaptive Guidance Adversarial Training} (AdaGAT) method, which incorporates MSE and RMSE loss functions with a stop-gradient operation. Our proposed MSE loss addresses the issue of excessively high accuracy in the learnable guide model. Additionally, the RMSE loss enhances the model’s sensitivity to subtle output differences.

% We identify the instability of adversarial robustness in the learnable model and its impact on the target model’s robustness. By examining the clean accuracy of the guiding model and the robustness of the target model, we interpret this issue as being due to excessively high clean accuracy in the guiding model.

\item To demonstrate that the {adaptive guidance adversarial training} (AdaGAT) method surpasses the robustness of other methods, we evaluate the effectiveness of our approach on three widely-used datasets: CIFAR-10~\cite{krizhevsky2009learning}, CIFAR-100~\cite{krizhevsky2009learning}, and Tiny-ImageNet~\cite{krizhevsky2009learning}, showing substantial improvements in model robustness across a wide range of adversarial attacks and baselines.

\end{itemize}

This paper is organized as follows: Sec.~\ref{Problem Definition And Motivation} provides a brief overview of the learnable adversarial training methods, highlights the issues encountered in adversarial training, and provides the motivation for our proposed approach. In Sec.~\ref{Proposed Method}, we present our proposed adversarial training method for a learnable architecture in detail. Sec.~\ref{Observations  and Analysis} outlines the adversarial training process applied to the target and guide models on the CIFAR-10 dataset~\cite{krizhevsky2009learning}, comparing our method with the baseline. Sec.~\ref{sec:experiment} concludes this work. 

%with a summary of findings and discusses potential improvements and applications of the proposed approach.
\section{Related work}

\subsection{Adversarial Attack}
%  Since their discovery fast gradient sign method (FGSM)~\cite{goodfellow2014explaining} stands as a prevalent technique for producing adversarial
% sample,
%  PGD~\cite{madry2017towards} utilized adaptive perturbations to update the attack direction. C$\&$W~\cite{carlini2017towards} consider optimizing the distance between adversarial examples and benign samples and proposed optimization-based attacks. combined the proposed method with two complementary adversarial attack methods to evaluate the robustness, which was called AutoAttack(AA)~\cite{croce2020minimally}~\cite{andriushchenko2020square}.

Since its discovery, the fast gradient sign method (FGSM)~\cite{goodfellow2014explaining} has shown that adversarial training is a foundational method to obtain a robust classifier from the perspective of min-max optimization for generating adversarial examples. Subsequently, the projected gradient descent (PGD)~\cite{madry2017towards} attack was introduced, enhancing FGSM by using adaptive perturbations to refine the attack direction. Furthermore, Carlini and Wagner (C$\&$W)~\cite{carlini2017towards} expanded on these ideas by optimizing the distance between adversarial and benign samples, thereby developing a family of optimization-based attacks. To enable even stronger attacks, AutoAttack (AA)~\cite{croce2020minimally} was recently proposed as a robust evaluation framework that combines the newly introduced method with two complementary adversarial attack techniques~\cite{andriushchenko2020square}, enabling a more comprehensive assessment of the robustness of the model. One of the effective methods for improving model robustness is adversarial training~\cite {rice2020overfitting}.

\subsection{Adversarial Training Methods}
\subsubsection{Adversarial Training with Single Models}
One effective defense method in adversarial settings is adversarial training (AT)~\cite{rice2020overfitting}, which directly trains deep neural networks (DNNs) on adversarial examples generated from projected gradient descent (PGD) attacks. Later, such as TRADES~\cite{zhang2019theoretically} highlighted an inherent trade-off between adversarial robustness and clean accuracy in existing AT methods. The adversarial weight perturbation (AWP)~\cite{wu2020adversarial} method addresses this trade-off by introducing perturbations to the model weights. Additionally, geometry-aware instance-reweighted adversarial training (GAIRAT)~\cite{zhang2020geometry} has shown improved robust accuracy under PGD attacks. In addition to adversarial training for a single model, some methods leverage large pre-trained models to achieve stronger robustness.

% Additionally, Cui et al. proposed learnable boundary-guided adversarial training (LBGAT), which uses a smaller guiding model to assist a larger target model. 
\subsubsection{Distillation-based Defense using Large Pre-Trained Models}
Despite several previous efforts, substantial progress in addressing adversarial robustness has remained limited. Recent advancements have largely stemmed from methods that leverage large pre-trained teacher models to improve robustness. For example, DGAD~\cite{park2025dynamic} utilizes the correctly classified output of a modified teacher model to train the student model. DARWIN~\cite{dong2024robust} generates intermediate adversarial samples to adjust the weights, better managing the trade-off in knowledge distillation and applying it to both targeted and untargeted phenomena. GACD~\cite{10018268} differentiates between the classifications of teacher and student models, utilizing a re-weighting strategy and latent representations to mitigate the teacher’s adversarial effects and improve the robustness of the student model. Although large pre-trained models can enhance the robustness of the target model, their teacher models require even stronger robustness and larger pre-trained models.  Some methods based on auxiliary models are not subject to this limitation, while also strengthening the target model’s robustness.
% \subsubsection{Distillation-Based Defense Using Large Pre-Trained Models}
% Despite several previous efforts, there has been limited substantial progress in addressing this robustness problem. Most impressive advancements in recent years come from those methods introducing extra large pre-train model.

% This means that improving robustness through large pre-trained teacher models comes at the expense of high memory usage, compared with standard training. This increased resource demand can negatively impact the experience of benign users and significantly reduce the adoption of adversarial training (AT) among real-world DNN application providers. A widely recognized solution to mitigate this issue is to use a trainable guide model, which enables smaller models to adaptively support larger models, particularly by training with a guiding model.

% DGAD~\cite{park2025dynamic} uses the modified teacher model's output, which is correctly classified, to train the student model. DARWIN~\cite{dong2024robust} uses intermediate adversarial samples to generate weights, which can better manage the trade-off in knowledge distillation and is applicable to both target and un-target phenomena. GACD~\cite{10018268} discriminates between the teacher model and student model classifications, using a re-weighting strategy to eliminate the teacher's adverse effects and enhance the student model's robustness.

\subsubsection{Adversarial Training with Auxiliary Models} Several approaches employ additional auxiliary models to support the training of the target model. LBGAT~\cite{cui2021learnable} employs a learnable guide model to generate clean outputs that support the adversarial training of the target model. In contrast, some auxiliary models explore adversarial generation behaviors during training. LAS-AT~\cite{jia2022adversarial} employs a trainable model to generate adversarial samples, enabling more effective adversarial training. 

% TAAT~\cite{kuang2024defense} employs a knowledge-guided approach based on global topology alignment and demonstrates that auxiliary models of varying sizes have been shown to considerably strengthen the target model's robustness.

However, due to the instability of certain auxiliary models during training, such as the guide model in~\cite{cui2021learnable}, they may hinder the target model from achieving optimal robustness. To enable auxiliary models to support the target model effectively to achieve higher robustness, we propose the Adaptive Guidance Adversarial Training (AdaGAT) method.
\section{Adaptive Guidance Adversarial Training}

\subsection{Problem Definition and Motivation}
\label{Problem Definition And Motivation}
%Empirically, our approach is inspired by and builds upon the previously proposed involvement of learnable guide models in the LBGAT method. 
We hypothesize that adjusting the guide model's performance towards the target model is crucial in improving robustness. To investigate this hypothesis, we monitor the states of both the guide and target models' performances throughout the adaptive adversarial training process. By examining the performance trends illustrated in Fig.~\ref{fig:combined}, we observe that maintaining a moderately low accuracy for the guide model can be beneficial for the robustness of the target model. Our step-by-step analysis reveals that this approach effectively addresses the identified issues and offers novel insights: the stop-gradient operation and a split additional loss ensure the guide model consistently aligns with the adversarial outputs of the target model, mitigating the risk of excessively rapid training of the guide model. Experimental results demonstrate that our novel \textit{adaptive guidance adversarial training} (AdaGAT) method achieves superior robustness in adversarial training, significantly outperforming baseline methods.
% \begin{figure}[h]
% \centering
% \vspace{0em}
% \includegraphics[width=1.00\linewidth]{notop_basediagram.pdf}
% \vspace{0em}
% \caption{Guide model's performance with and without backpropagation. The \textbf{gray line} indicates the accuracy of guide (with {\color{gray}$\CIRCLE$}) and target (with {\large \color{gray}$\smblksquare$}) models when the guide model does not participate in backpropagation, while \textbf{red line} represents the accuracy of guide (with {\color{red}$\CIRCLE$}) and target (with {\large \color{red}$\smblksquare$}) when the guide model participates in backpropagation. The results show the robustness accuracy (evaluated under a PGD-20 attack) on CIFAR-10.}
% \textbf{} 
% \label{fig:LBGAT and NOP}
% \vspace{0em}
% \end{figure}

To investigate how learnable guide models influence the target model during adversarial training, we evaluate the clean accuracy of the guide model and the robustness of the target model under two settings: with and without backpropagation through the learnable guide model. Fig.~\ref{fig:combined} (a) reveals that the learnable guide model (with backpropagation) improves the target model's robustness under adversarial attacks better than the guide model (without backpropagation). Remarkably, this performance is observed despite the learnable guide model (with backpropagation) having relatively lower clean accuracy compared to the learnable guide model (without backpropagation). We, therefore, hypothesize that the learnable properties of the guide model (i.e., adaptively co-training the guide model along with the target model) can improve the robustness of the target model. 

From Fig.~\ref{fig:combined} (a), we observe that the target model causes the lowering of the guide model's clean accuracy (via backpropagation). Conversely, a lower clean accuracy of the guide model positively impacts the robustness of the target model. Based on this observation, we hypothesize that improving or increasing the strength of the backpropagation to the guide model while decreasing its clean accuracy could lead to enhanced robustness of the target model. This required effective synchronization for dynamic guide-target co-training. Thus, we propose an additional function to align the clean accuracy of the guide model with the robust accuracy of the target model. We also argue that a single loss function of previous work~\cite{cui2021learnable} may be insufficient to constrain this pivotal guide model adequately in the dynamic guide-target co-training. Thus, we propose incorporating adaptive loss to ensure that the learnable guide model closely aligns with the target model, while the target model does not actively participate in this alignment process, thereby further enhancing the latter's robustness against adversarial attacks.  

% These previous regularization methods only focus on equally minimizing the discrepancy between predictions for clean and adversarial output but
% lack control guide model training.

% Motivated by this finding, we introduce a new regularizer that more effectively constrains the output of the target model.

% To address this, we introduce an additional function to further constrain the target model’s output, enhancing alignment between the guiding and target models and thus contributing to a more robust overall training framework.

\subsection{Proposed Method}
\label{Proposed Method}
\subsubsection{Notations}
Let \( x \in \mathbb{R}^d \) be an image and \(y \in \{1,2,\ldots, K\} \) be a class label in an image classification problem with a dataset \( D = \{(x_i, y_i) \mid i = 1,2,\ldots, n \} \). Let  \( f_\theta:\mathbb{R}^d \rightarrow \mathbb{R}^K \) be a DNN classifier parametrized by learnable parameters $\theta$. Thus, we consider a guide model as a DNN classifier \(f_{\theta_g} \) that learns to classify images such that \( f_{\theta_g}(x) \) represents the output logits of the guide model that has parameters \( \theta_g \). Similarly, \( f_{\theta_t}(x + \delta) \) with target model parameters \( \theta_t \) represents the adversarial output of a target model, where \( \delta \in \mathbb{R}^d \) represents the adversarial perturbation which is typically bounded by \( \|\delta\|_{\infty} \leq \epsilon \), where \( \epsilon \) is a small constant that determines the maximum allowed perturbation.

% In the context of clean training for guide, for a classification task with $k \geq 2$ as the number of classes, given a dataset $\{D = {(x_{i}, y_{i}) | (1 \leq i \leq 1 )}\}$ with $x_i \in X$, $y_i \in Y (Y \in R^d)$ respectively denoting a natural sample and its supervised label, guide model aims to learn a classifier $f_{g}$, such that $f_{g}(x_i, \theta_t)$ represents the output logits with guide model parameters. For the target model's adversarial training, let $f_{t}(x_{i}+ \delta,\theta_t)$ denote the adversarial output of the target model. Similarly, we define $f_{t}(x_i+ \delta, \theta_t)$ to represent the adversarial output logits with the target model parameters $\theta_t$. This notation, $\delta$, represents the $l_{\infty}$-norm bound on the adversarial perturbation, where is typically the maximum allowed value for the perturbation.

\subsubsection{Preliminaries}
The method LBGAT can be formalized as two optimizations with constraints between the guide model and the target model. The guide model is trained on ground-truth (clean) data using cross-entropy during the training phase. Additionally, mean squared error (MSE) is applied during adversarial training, optimizing both the guide model and the target model simultaneously. This dual-purpose approach not only strengthens the adversarial training of the target model but also enhances the optimization of the guide model. The process of adversarial training can be defined as:
\begin{equation}
\begin{split}
    \ell(x,y) &= \mathop{\min}_{\theta_t,\theta_g} \big\{ \mathcal{L}_{\text{CE}}\big(f_{\theta_g}(x), y\big)  \\
    &\quad + \mathcal{L}_{\text{MSE}}\big(f_{\theta_t}(x + \delta), f_{\theta_g}(x)\big)\big\},
\end{split}
\label{eq:advloss_function}
\end{equation}
where the cross-entropy loss $ \mathcal{L}_{\text{CE}}(\cdot,\cdot)$ is adopted to scale the performance of the guide classifier $f_{g}$. Additionally, the mean squared error loss $ \mathcal{L}_{\text{MSE}}(\cdot, \cdot)$ is employed to optimize the performance of both the target model and the guide model. In Eq.~\eqref{eq:advloss_function}, both guide and target models share an MSE loss as:
\begin{equation}
\begin{split}
    \mathcal{L}_{share} & = \mathcal{L}_{\text{MSE}}\big(f_{\theta_t}(x + \delta), f_{\theta_g}(x)\big),
\end{split}
\label{eq:shared_loss}
\end{equation}
which allows equal backpropagation to both guide and target models.

\subsubsection{Adaptive Guide Models in Adversarial Training}
We hypothesize that relying solely on a single MSE-based equal backpropagation of both guide and target poses limitations in achieving a highly robust state of the target model. To address this issue, our research focuses on tuning the guide model with an additional loss function to align its clean output with the adversarial state of the target model. From a contribution perspective, our approach incorporates a new regularization term to enhance the target model's adversarial robustness.

The incorporation of this regularization term is a well-considered decision. Our method adopts an additional adaptive MSE function compared to the LBGAT formulation in Eq.~\eqref{eq:advloss_function} for two main reasons. First, previous work has shown that MSE effectively enhances robustness~\cite{cui2021learnable}. Second, an MSE loss would maintain the consistency of the loss function, as other loss functions would make it challenging to intuitively adjust the adversarial training states of both models. % Therefore, we propose that substituting KL divergence with MSE strikes a better balance between robustness and clean accuracy. 
Therefore, our loss function that performs the AdaGAT co-training is formulated as follows:
\begin{equation}
\begin{split}
    \mathcal{L}_{\text{AdaGAT-MSE}} &= \mathop{\min}_{\theta_g} \Bigl\{    \mathcal{L}_{\text{CE}}\left(f_{\theta_g}(x), y\right)  \\
    &\quad + \mathcal{L}_{share} + \lambda \,\mathcal{L}_{\text{adaMSE}} \left( \overline{f_{\theta_t}(x + \delta)}, f_{\theta_g}(x)\right) \Bigl\},
\end{split}
\label{eq:advloss_function}
\end{equation}
where $\overline{f_{\theta_t}(x + \delta)}$ indicates that the target model does not perform back-propagation, the hyperparameter $\lambda$ indicates that it controls the penalty strength of regularization, $ \mathcal{L}_{share}$ indicates the shared training process in which both the guide model and the target model participate collaboratively. The target model's involvement is limited solely to this shared training process for optimizing the target model parameters:
\begin{equation}
\begin{split}
    \mathop{\min}_{\theta_t} \big\{ \mathcal{L}_{share}\big\},
\end{split}
\label{eq:Target_loss}
\end{equation}
%
% Building upon prior research, we aim to incorporate an additional loss term to tune the guide model's clean output in order to more effectively approximate the adversarial outputs of the target model. Furthermore, we strive to implement adaptive guidance that closely aligns with the variability of adversarial outputs while maintaining consistency with the loss functions employed in previous methods. To address these requirements, we propose a straightforward yet distinctive $\mathcal{L}_{\text{MSE}}$ loss, derived from the clean output corresponding to the adversarial output, guided by the following key considerations:

% First, clean accuracy generally exceeds robust accuracy by a large margin, and the lightweight model is easier to optimize, making them more suitable as target models. 

% Second, when performing adversarial training on datasets with numerous classes, the robust accuracy for some classes may drop excessively, even approaching zero, while the guide model’s clean accuracy remains high. This discrepancy makes it ineffective for guidance adversarial training, whereas using a more adaptive clean accuracy avoids this issue. In other words, we believe that if the guide model's clean accuracy better aligns with the adversarial outputs of the target model, it will be beneficial for the robustness of the target model. 

Initially, similar auxiliary functions (such as the use of MSE functions) allow the intuitive adjustment of hyperparameters according to different scenarios. That is, integrating the loss function effectively enhances the target model's robustness despite the guide models' clean accuracy deteriorating. 

% Additionally, if the guide model's clean accuracy significantly exceeds the target model's robust accuracy, it could potentially affect the adversarial training for the target model. To address this issue, we reduce the clean accuracy of the guide model by aligning it more closely with the robust accuracy. This alignment allows the clean outputs of the guide model to more effectively guide the adversarial outputs of the target model, thereby enhancing its robustness. Conversely, when the target model demonstrates high robustness, this situation will appropriately reduce its influence on the guide model. To demonstrate the effectiveness of our method, we leverage the analysis results shown in Fig.~\ref{fig:combined} (b) to introduce accuracy alignment between the guide model and the target model. Additionally, we adjust the parameter~$\lambda$ to determine the influence of the target model's adversarial output on the guide model, which enables the target model to achieve optimal robustness.

When the difference between two outputs is large, the Mean Squared Error (MSE) function amplifies the discrepancy. However, as outputs become closer, particularly when the difference is less than 1, the MSE becomes less sensitive, which may hinder the model's ability to further minimize small but important differences. The root mean square error (RMSE), as a variant of MSE, has been shown to emphasize large errors while simultaneously improving sensitivity to small differences~\cite{karunasingha2022root}.  Compared to the MSE mentioned above, the RMSE is more sensitive to small output differences, as the square-root operation tends to amplify errors smaller than one. Motivated by this observation, we propose a loss function combining the advantages of both MSE and RMSE.

\begin{equation}
\begin{array}{rl}
 \mathcal{L}_{\text{adaRMSE}} \left( \overline{f_{\theta_t}(x + \delta)}, f_{\theta_g}(x)\right) = & \\
 & \hspace{-2cm}  \sqrt{\mathcal{L}_{\text{adaMSE}} \left( \overline{f_{\theta_t}(x + \delta)}, f_{\theta_g}(x)\right)},
\end{array}
\label{eq:RMSEdefin_function}
\end{equation}

where $\sqrt{.}$ represents the square root.

\begin{equation}
\begin{split}
    \mathcal{L}_{\text{AdaGAT-RMSE}} &= \mathop{\min}_{\theta_g} \Bigl\{    \mathcal{L}_{\text{CE}}\left(f_{\theta_g}(x), y\right) + \mathcal{L}_{share} \\ &\quad  + \lambda \,\mathcal{L}_{\text{adaRMSE}} \left( \overline{f_{\theta_t}(x + \delta)}, f_{\theta_g}(x)\right)\Bigl\},
\end{split}
\label{eq:RMSEloss_function}
\end{equation}

From a mathematical perspective, RMSE geometrically represents the Euclidean distance between the clean model output and the adversarial target output, which intuitively measures their disparity. We compare proposed methods, and the robust accuracy results are shown in Sec.~\ref{sec:experiment}. It can be observed that compared to AdaGAT-MSE, the proposed AdaGAT-RMSE achieves better adversarial robustness across all attack scenarios on both the CIFAR-10 and CIFAR-100 datasets. Our method AdaGAT is summarized in Algorithm~\ref{alg:1}.

\begin{algorithm}[h!]
%\small
   \caption{Adaptive Guidance Adversarial Training}
   \label{alg:1}
\begin{algorithmic}[1]
   \STATE {\bfseries Input:} Guide model $f_g$, target model $f_t$, training data $D$, mini-batch $X$, batch size $m$,  $\lambda$ is a hyper-parameter.
   \STATE {\bfseries Output:} Guide model $\theta_g$, target model $\theta_t$.
   \REPEAT
   \STATE Read mini-batch $X$ from a training set $D$.
   \STATE Get adversarial examples $X^{adv} $ by an adversarial attack on input $X$.
   \STATE Obtain clean output $f_{g}(x)$ for $x\in X $ from guide model.
   \STATE Obtain adversarial output $f_{t}(x^{adv})$ from target model.
   \STATE Construct adversarial output and clean output with the same shared loss as per Eq.~\eqref{eq:shared_loss}.
   \STATE Optimize guide model $f_g$ as per Eq.~\eqref{eq:advloss_function} or Eq.~\eqref{eq:RMSEloss_function} .
   \STATE Optimize target model $f_t$ as per Eq.~\eqref{eq:Target_loss}.
   \UNTIL{maximum training epoch reached}
\end{algorithmic}
\end{algorithm}

% \subsubsection{Adaptive Guide Models in Adversarial Training}

% Our method AdaGAT is summarized in Algorithm~\ref{alg:1}.

% \begin{algorithm}[h!]
% %\small
%    \caption{Adaptive Guidance Adversarial Training}
%    \label{alg:1}
% \begin{algorithmic}[1]
%    \STATE {\bfseries Input:} Guide model $f_g$, target model $f_t$, training data $D$, mini-batch $X$, batch size $m$,  $\lambda$ is a hyper-parameter.
%    \STATE {\bfseries Output:} Guide model $\theta_g$, target model $\theta_t$.
%    \REPEAT
%    \STATE Read mini-batch $X$ from a training set $D$.
%    \STATE Get adversarial examples $X^{adv} $ by an adversarial attack on input $X$.
%    \STATE Obtain clean output $f_{g}(x)$ for $x\in X $ from guide model.
%    \STATE Obtain adversarial output $f_{t}(x)$ from target model.
%    \STATE Construct adversarial output and clean output with the same shared loss as per Eq.~\eqref{eq:shared_loss}.
%    \STATE Optimize guide model $f_g$ as per Eq.~\eqref{eq:advloss_function}.
%    \STATE Optimize target model $f_t$ as per Eq.~\eqref{eq:Target_loss}.
%    \UNTIL{maximum training epoch reached}
% \end{algorithmic}
% \end{algorithm}

\section{Observations and Analysis}
\label{Observations  and Analysis}
\begin{figure}[t]
    \centering
    \subfloat[\label{fig:a}]{\includegraphics[width=0.49\linewidth]{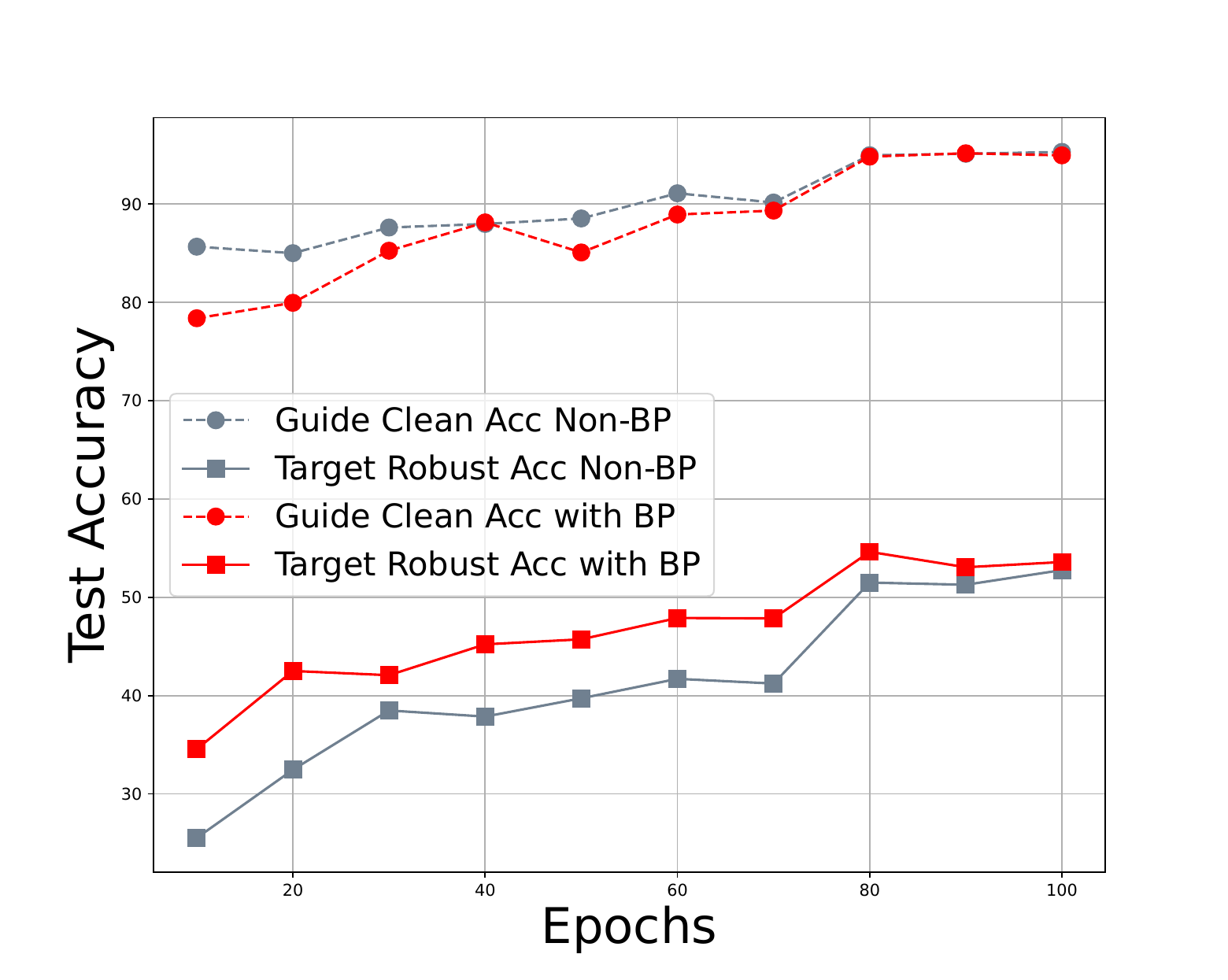}} \hfill
    \subfloat[\label{fig:b}]{\includegraphics[width=0.49\linewidth]{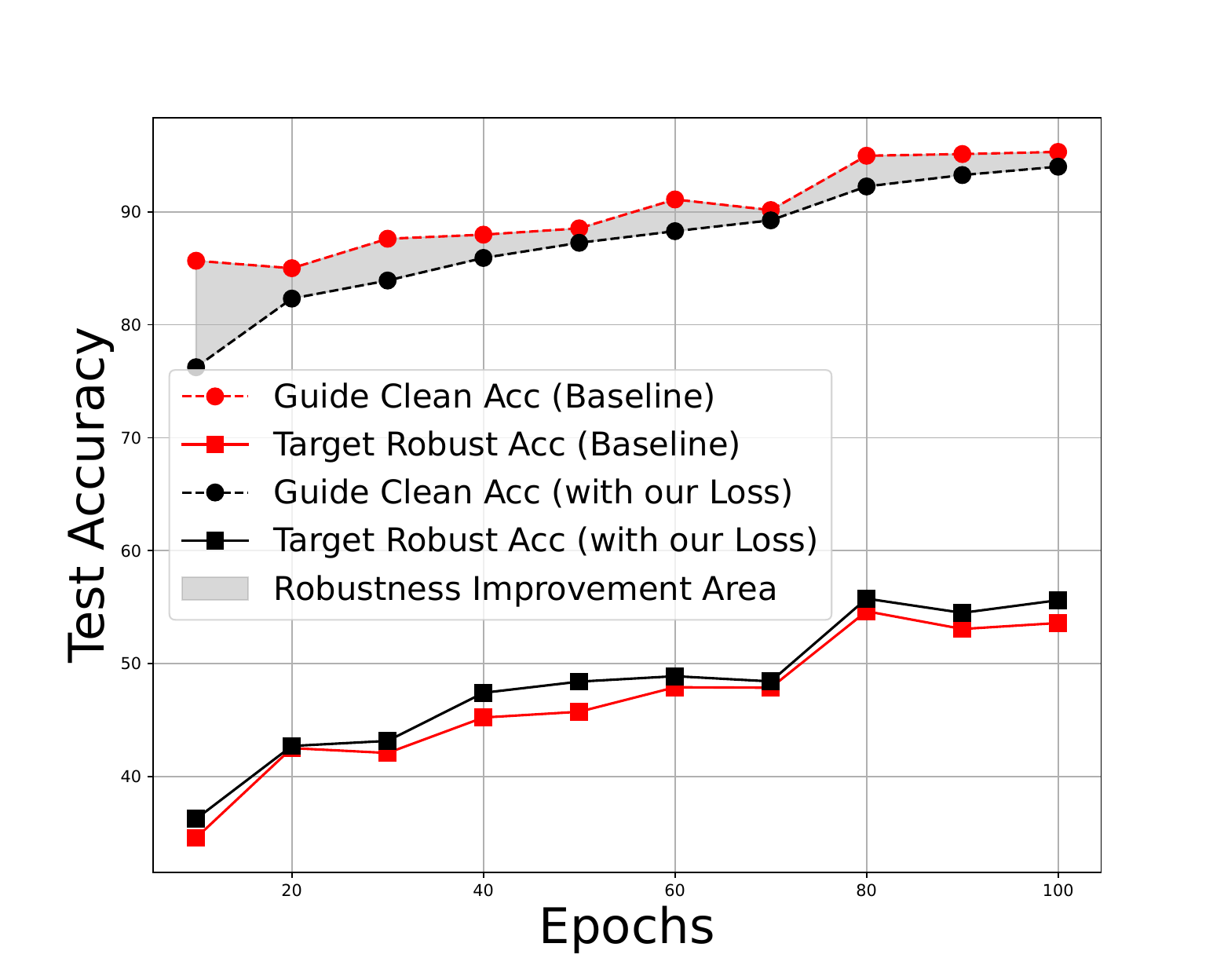}}
    \caption{Comparison of the guide model's performance with and without backpropagation, and the effect of applying our method.  
    \textbf{(a)} The \textbf{gray line} indicates the accuracy of the guide (with {\color{lightgray}$\CIRCLE$}) and target (with {\large \color{lightgray}$\smblksquare$}) models when the guide model does not participate in backpropagation, while the \textbf{red line} indicates the accuracy of the guide (with {\color{red}$\CIRCLE$}) and target (with {\large \color{red}$\smblksquare$}) models when the guide model participates in backpropagation.  
    \textbf{(b)} The \textbf{red line} has the same meaning as mentioned above and indicates the accuracy of the baseline method, while the \textbf{black line} represents the accuracy of guide (with {\color{black}$\CIRCLE$}) and target (with {\large \color{black}$\smblksquare$}) models when the models follow our loss. The shaded \colorbox{gray!70}{\phantom{ }} region indicates intervals where the guide model strengthens the robustness of the target model. The above results show robust accuracy (evaluated under a PGD-20 attack) on CIFAR-10.}
    \label{fig:combined}
    % \label{fig:comparison}
\end{figure}

We analyzed the performances of guide and target models with a single MSE (the baseline loss) and our adaptive guidance loss function over 100 epochs and under a PGD-20 attack. Specifically, every 10 epochs, we record the performance of both the target model and the guide model. To make the guide model’s impact on improving the robustness of the target model more apparent, we adopt a scaling factor into the loss function. Fig.~\ref{fig:combined} (b) shows that despite the clean accuracy of the guide model dropping below the single MSE (the baseline loss), the robust accuracy of the target model improved against the robustness achieved by the baseline loss. 

% \begin{figure}[h!]
% \centering
% \vspace{0em}
% \includegraphics[width=1.10\linewidth]{notop_ourdiagram.pdf}
% \vspace{0em}
% \caption{Comparison of the model's accuracy before and after applying our method. The \textbf{red line} indicates the accuracy of guide (with {\color{red}$\CIRCLE$}) and target (with {\large \color{red}$\smblksquare$}) models when the models follow LBGAT loss, while \textbf{black line} represents the accuracy of guide (with {\color{black}$\CIRCLE$}) and target (with {\large \color{black}$\smblksquare$}) models when the models follow our loss. The shaded \colorbox{gray!35}{\phantom{ }} region indicates intervals where the guide model improves the robustness of the target model. The results show the robustness accuracy (evaluated under a PGD-20 attack) on CIFAR-10.}
% \label{fig:30_adv and Entire}
% \vspace{0em}
% \end{figure}

Theoretically, the robustness of the target model is closely related to the clean accuracy of the guide model. It is evident that an appropriately low accuracy of the guide model contributes to improving the robustness of the target model. In ~Fig.~\ref{fig:combined} (b), we illustrate the range of the guide model's performance within which it effectively enhances the robustness of the target model. However, experimental results indicate that if the guide model's performance exceeds this range, the robustness of the target model is adversely affected, as shown in Table~\ref{table:ablation} in Sec.~\ref{sec:experiment}. Therefore, our approach aims to identify the most suitable performance level of the guide model (indicated by the black dashed line in Fig.~\ref{fig:combined} (b)) that maximizes the robustness of the target model (represented by the solid black line in Fig.~\ref{fig:combined} (b)).

\section{Experiment}
\label{sec:experiment}

\subsection{Default Training Setting}
\subsubsection{Experimental Setups}

To highlight the effectiveness of our proposed method, we adopt the same adversarial training architecture and hyperparameter settings as LBGAT~\cite{cui2021learnable}, including the perturbation configurations and the number of iteration steps. Specifically, both our method and LBGAT utilize ResNet-18~\cite{he2016deep} as the guide model, under which the target model achieves its best robustness performance, and WideResNet-34-10 as the target model.

% We use the same adversarial training architecture and parameters as LBGAT~\cite{cui2021learnable}, including perturbation settings and iteration steps. The guide models in our experiment and in LBGAT were implemented with ResNet-18~\cite{he2016deep}, and the target models were WideResNet-34-10~\cite{he2016deep}. 
\subsubsection{Datasets}
We conduct experiments using three benchmark datasets: CIFAR-10~\cite{krizhevsky2009learning}, CIFAR-100~\cite{krizhevsky2009learning}, and Tiny-ImageNet~\cite{krizhevsky2009learning}, which are widely employed to evaluate adversarial robustness. CIFAR-10 and CIFAR-100 each contain 50,000 training images and 10,000 test images, with dimensions of 32 × 32 × 3. Regarding the number of classes, CIFAR-10 includes 10 classes, while CIFAR-100 contains 100 classes. Tiny-ImageNet is a subset of the ImageNet dataset, encompassing 200 classes, with each class containing 600 images. For consistency, the image size in Tiny-ImageNet is also set to 32 × 32 × 3.
% We trained all the models over 100 epochs, utilizing a batch size of 128.  We use an SGD momentum optimizer with a weight decay of $5 \times 10^{-4}$. The learning rate was initially set to $0.1$, and the learning rate decay was set to a factor of $0.1$ to operate at the 75th and 90th epochs. Additionally, a fixed warm-up was applied in the 1st epoch with a learning rate of 0.02, followed by 0.1 for the remaining epochs.. We adopt the Fast Gradient Method (FGSM), PGD attack with 10 steps (PGD-10), 20 steps (PGD-20), and 50 steps (PGD-50), the C$\&$W attack with 20 steps (C$\&$W), and AutoAttack (AA) to evaluate the performance of the trained models.

% \subsubsection{Datasets}
% We conduct experiments using three benchmark datasets: CIFAR-10~\cite{krizhevsky2009learning}, CIFAR-100~\cite{krizhevsky2009learning}, and Tiny-ImageNet~\cite{krizhevsky2009learning}, which are widely employed to evaluate adversarial robustness. CIFAR-10 and CIFAR-100 each contain 50,000 training images and 10,000 test images, with dimensions of 32 × 32 × 3. Regarding the number of classes, CIFAR-10 includes 10 classes, while CIFAR-100 contains 100 classes. Tiny-ImageNet is a subset of the ImageNet dataset, encompassing 200 classes, with each class containing 600 images. For consistency, the image size in Tiny-ImageNet is also set to 32 × 32 × 3.

\subsection{Detailed Hyper-Parameter Settings}
Our method's hyperparameter settings follow the previously proposed LBGAT method. For the additional loss function introduced in our method, we adopt a regularization function combined with the MSE loss function shown in Eq.~\eqref{eq:shared_loss}. A key hyperparameter in our proposed method is \(\lambda\), which balances our additional loss and the shared loss as shown in Eq~\eqref{eq:advloss_function}. The results of our experiments with different values of \(\lambda\) are shown in Table~\ref{table:ablation}. When \(\lambda = 2.5\), the target model achieves the best robustness accuracy across all attack scenarios. The detailed analysis and explanation of this phenomenon can be found in Section~\ref{Parameter}.

% Specifically, under the AutoAttack (AA), our method achieves an accuracy of approximately 53.59\%. These results demonstrate that the choice of \(\lambda\) plays a critical role in balancing clean accuracy and robustness, which we further analyze in the ablation study.

\subsection{Performance Comparison}
To validate the performance of the proposed method, we conduct a comparison of the proposed method with several state-of-the-art adversarial training methods, i.e., 
PGD-AT ~\cite{rice2020overfitting}, 
TRADES ~\cite{zhang2019theoretically}, 
MART ~\cite{wang2019improving}, 
FAT  ~\cite{zhang2020attacks}, 
GAIRAT ~\cite{zhang2020geometry}, and
%AWP  ~\cite{wu2020adversarial}, 
LBGAT ~\cite{cui2021learnable}. 
Note that we adopt the training settings reported in their original works to train these AT models.

\subsubsection{Comparison Results on CIFAR-10}
As for the CIFAR-10 dataset, we use WideResNet-30-10 as the backbone. The results are shown in Table~\ref{table:cifar10}.
It can be observed that compared with the advanced defense method, the proposed method achieves better model robustness under the AA and C\%W attacks. % and requires much less trainingtime. 
%Compared with other advanced fast adversarial training methods, the proposed method achieves the best model robustness under all attack scenarios for the best and last checkpoints. 
Specifically, under the AA attack, the baseline adversarial training methods achieve robustness accuracy of approximately 52.23\%, while our proposed AdaGAT-MSE improves this performance to around 53.59\%, offering an increase of 1.36\%. AdaGAT-RMSE achieves a robustness accuracy of 53.87\%.

\subsubsection{Comparison Results on CIFAR-100}
As for CIFAR-100, WideResNet-30-10 is used as the backbone network. The result is shown in Table~\ref{table:cifar100}. Compared with PGD-AT, the proposed method achieves better model robustness under most attack scenarios. Even under the $C\&$W attack, the baseline method obtains an accuracy of about 28.72$\%$, while AdaGAT-MSE achieves an accuracy of about 29.24$\%$, and AdaGAT-RMSE achieves an accuracy of 29.37$\%$.

\begin{table}[h!]
\centering
\renewcommand{\arraystretch}{1.2}
\setlength{\tabcolsep}{4pt}
\caption{Robustness (\%) evaluated on the CIFAR-10 dataset with the WRN34-10 architecture.
% Asterisk (*) indicates that the results were excerpted from the papers and † indicates our reproduction 
}

\label{tb:cifar10}

%\vspace{0mm}

\begin{tabular}{lccccc}
\toprule

Method    & PGD-10         & PGD-20         & PGD-50         & C\&W           & AA             \\ \midrule 
PGD-AT ~\cite{rice2020overfitting}   & 56.07          & 55.08          & 54.88          & 53.91          & 51.69          \\ 
TRADES ~\cite{zhang2019theoretically} & 56.75          & 56.1           & 55.9           & 53.87          & 53.40          \\ 
MART ~\cite{wang2019improving}  & 58.98          & 58.56          & 58.06          & 54.58          & 51.10          \\ 
FAT  ~\cite{zhang2020attacks} & 50.31          & 49.86          & 48.79          & 48.65          & 47.48          \\ 
GAIRAT ~\cite{zhang2020geometry}  & \textbf{60.64}          & \textbf{59.54}          & \textbf{58.74}          & 45.57          & 40.30          \\ 
% AWP  ~\cite{wu2020adversarial}            & 85.57          & 58.92          & 58.13          & 57.92          & 56.03          & 53.90          \\
LBGAT ~\cite{cui2021learnable} (baseline)  &  56.25      &        54.66        &           54.3     &          54.29      & 52.23 
\\  

\rowcolor{LightCyan} \textbf{AdaGAT-MSE (ours)}  &  57.63        & 56.52       & 56.11        &  {55.52}      & {53.59}          \\

\rowcolor{LightCyan} \textbf{AdaGAT-RMSE (ours)}  &  58.17        & 57.10       & {56.78}        &  \textbf{55.57}      & \textbf{53.87}          \\

% Method           & Clean          & PGD-10         & PGD-20         & PGD-50         & C\&W           & AA             \\ \midrule 
% PGD-AT ~\cite{rice2020overfitting}         & 85.17          & 56.07          & 55.08          & 54.88          & 53.91          & 51.69          \\ 
% TRADES ~\cite{zhang2019theoretically}          & 85.72          & 56.75          & 56.1           & 55.9           & 53.87          & 53.40          \\ 
% MART ~\cite{wang2019improving}            & 84.17          & 58.98          & 58.56          & 58.06          & 54.58          & 51.10          \\ 
% FAT  ~\cite{zhang2020attacks}            & 87.97     & 50.31          & 49.86          & 48.79          & 48.65          & 47.48          \\ 
% GAIRAT ~\cite{zhang2020geometry}           & 86.30          & 60.64          & 59.54          & 58.74          & 45.57          & 40.30          \\ 
% % AWP  ~\cite{wu2020adversarial}            & 85.57          & 58.92          & 58.13          & 57.92          & 56.03          & 53.90          \\
% LBGAT ~\cite{cui2021learnable} (baseline)          & 88.22          &          56.25      &        54.66        &           54.3     &          54.29      & 52.23 
% \\  

% \textbf{AdaGAT (ours)} & 86.05         &  \textbf{57.63}        & \textbf{56.52}       & \textbf{56.11}        &  \textbf{55.52}      & \textbf{53.59}          \\

\bottomrule
\end{tabular}

%\vspace{0mm}
\label{table:cifar10}
\end{table}

\begin{table}[h!]
\renewcommand{\arraystretch}{1.2}
\setlength{\tabcolsep}{4pt}
\centering
\caption{Robustness (\%) evaluated on the CIFAR-100 dataset with the WRN34-10 architecture.  
% Comparisons with state-of-the-art AT methods on the CIFAR100 database using WRN34-10. 
% Numbers represent percentage.  Number in bold indicates the best. 
%$\uparrow$ and $\downarrow$ indicate the performance change after the combination with our framework. 
% `a $\uparrow$ b' denotes that the accuracy is `a' and the improvement over the corresponding base model is `b'. `$\uparrow$' means `increase' and `$\downarrow$' means `decrease'. 
}
 \label{tb:cifar100}
%\vspace{0mm}
\begin{tabular}{lccccc}
\toprule
Method     & PGD-10         & PGD-20         & PGD-50         & C\&W             & AA             \\  \midrule
TRADES~\cite{zhang2019theoretically}  & 29.20          & 28.66          & 28.56          & 27.05          & 25.94          \\ 
SAT~\cite{sitawarin2021sat}  &     28.10           &       27.17         &  26.76              &    27.32            & 24.57          \\ 
LBGAT~\cite{cui2021learnable} (baseline)  &      32.05          &        30.77        &             30.42   &     28.72          & 27.16          \\

\rowcolor{LightCyan} \textbf{AdaGAT-MSE (ours)}  & {32.50}  & {31.59}   & {31.31}  & {29.24}  &  {27.69} \\
\rowcolor{LightCyan} \textbf{AdaGAT-RMSE (ours)}  & \textbf{32.63}  & \textbf{31.63}   & \textbf{31.35}  & \textbf{29.37}  &  \textbf{27.79}

 \\ \bottomrule
\end{tabular}
\label{table:cifar100}
%\vspace{0mm}
\end{table}

\subsubsection{Comparison Results on Tiny ImageNet}

\begin{table}[H]
\renewcommand{\arraystretch}{1.2}
\setlength{\tabcolsep}{12pt}
\centering
\caption{ Robustness (\%) on the (32x32) Tiny ImageNet dataset using WRN34-10, with WideResnet as the guide model for both LBGAT and our proposed methods. The best performance values are highlighted in bold.
% Comparisons with state-of-the-art AT methods on the Tiny Imagenet database using PreActResNet18. Number in bold indicates the best. 
%$\uparrow$ and $\downarrow$ indicate the performance change after the combination with our framework. 
% `a $\uparrow$ b' denotes that the accuracy is `a' and the improvement over the corresponding base model is `b'. `$\uparrow$' means `increase' and `$\downarrow$' means `decrease'. 
}
 \label{tb:tiny}
 %\vspace{-3mm}
\begin{tabular}{lccc}
\toprule 
Method     &FGSM      & PGD-10         & C\&W               \\ \midrule
TRADES ~\cite{zhang2019theoretically}  & 9.69  &5.01 &  4.63  \\ 
LBGAT~\cite{cui2021learnable} (baseline)   & 15.85      &  8.88   & 6.75  \\ 
\rowcolor{LightCyan} \textbf{AdaGAT-RMSE (ours)} & {15.86} & {9.02} & {7.18} \\
\rowcolor{LightCyan} \textbf{AdaGAT-MSE (ours)} & \textbf{16.04} & \textbf{9.25} & \textbf{7.21}  \\

% Method     & Clean    &FGSM      & PGD-10         & C\&W               \\ \midrule
% TRADES ~\cite{zhang2019theoretically}  & 31.01  &9.69  &5.01 &  4.63                 \\ 
% LBGAT~\cite{cui2021learnable} (baseline)    & 39.77    &15.85      &  8.88         & 6.75                   \\  

% \textbf{AdaGAT (ours)}  & 37.64 & \textbf{16.04} & \textbf{9.25} & \textbf{7.21}  \\ 
\bottomrule
\end{tabular}
\label{table:Tiny-imagenet34-10}
% \vspace{0mm}
\end{table}

As for TinyImageNet, following the previous works, WideResNet-30-10 is used as the target model and guide model. The result is shown in Table \ref{table:Tiny-imagenet34-10}. Compared with PGD-AT, our approach demonstrates superior adversarial robustness across all attack scenarios. Furthermore, under AA attack, LBGAT achieves an accuracy of about 6.75\%, for our method, AdaGAT-MSE, it achieves an accuracy of about 7.21\%. 

It is worth noting that we crop the original 64×64 images to 32×32 during training, which may introduce fluctuations in performance and cause discrepancies between the results of AdaGAT-MSE and AdaGAT-RMSE on TinyImageNet compared to other datasets. % Additionally, the clean accuracy achievedd at \( 37.64\% \).

\subsection{Ablation Study on the Choice of Parameter \(\lambda\)}
\label{Parameter}
To analyze the impact of the parameter \(\lambda\), we conduct experiments with values of \(\lambda\) set to 1, 2, 2.5, and 3. An example of the impact of \(\lambda\) on the CIFAR-10 dataset is summarized in Table~\ref{table:ablation}. Table~\ref{table:ablation} presents the model's performance across clean accuracy and robustness metrics, including PGD-10, PGD-20, PGD-50, C\&W, and AA attacks. For our setting, the parameter \(\lambda = 2.5\) achieves the optimal trade-off across all datasets and experiments, providing robust performance across all attack types while maintaining reasonable clean accuracy. Therefore, we select \(\lambda = 2.5\) as the optimal value for our proposed method. It is worth noting that excessively increasing $\lambda$ compromises the effectiveness of the guiding model. Specifically, an overly large $\lambda$ forces the guide model to approximate the target model too closely, which diminishes the guide model's clean accuracy and undermines its ability to provide effective, reliable guidance during training.

\subsection{Adversarial Loss Landscape Visualization and Analysis}

\begin{figure}[htbp]
    \centering
    \subfloat[Baseline Method]{\includegraphics[width=0.3\textwidth]{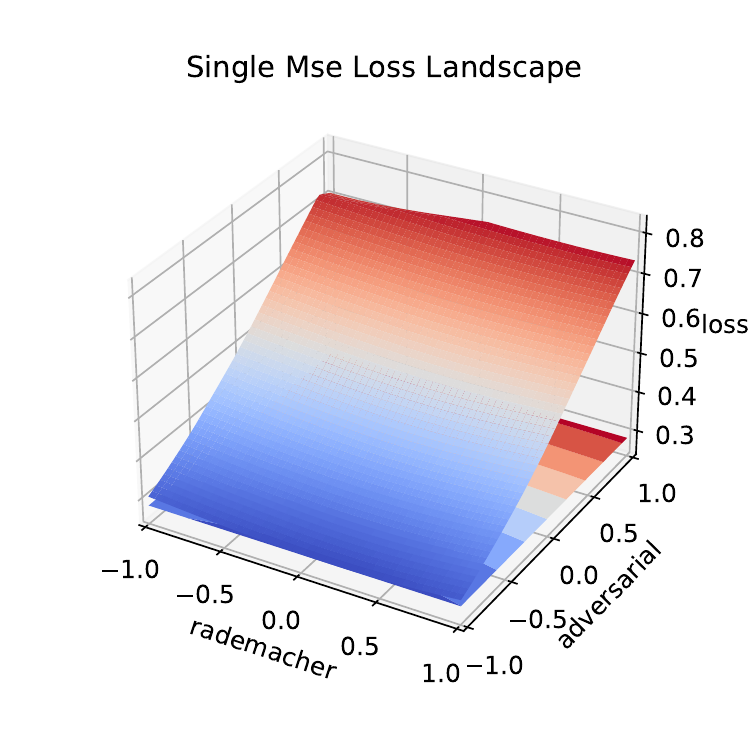}\label{fig:landscape_setup}}
    \hfill
    \subfloat[AdaGAT-MSE(ours)]{\includegraphics[width=0.3\textwidth]{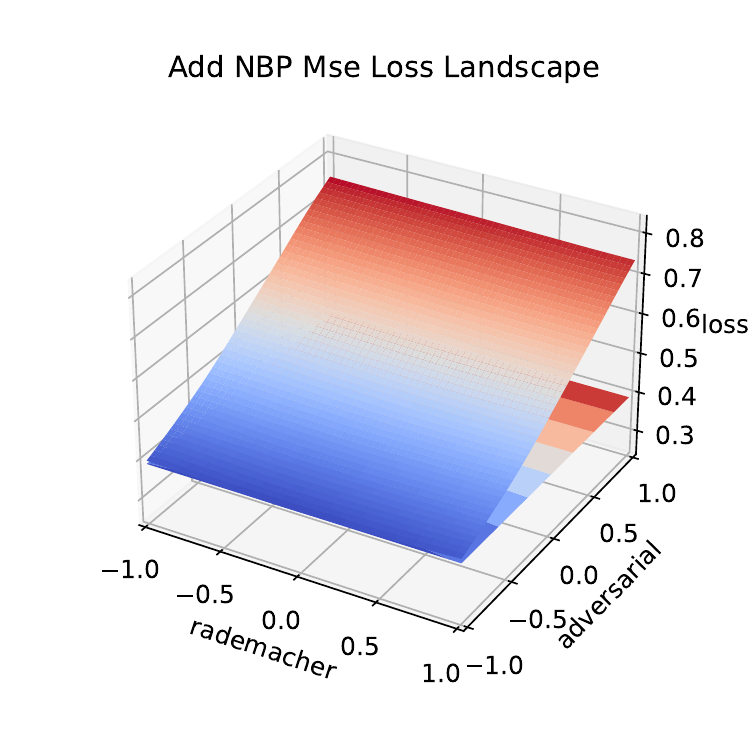}}
    \hfill
    \subfloat[AdaGAT-RMSE(ours)]{\includegraphics[width=0.3\textwidth]{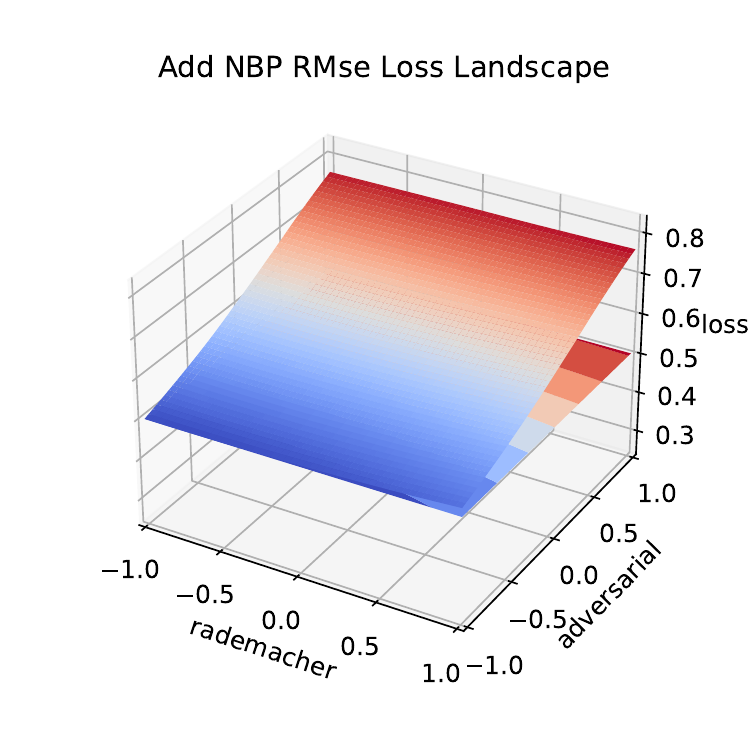}\label{fig:landscape_results}}
    \caption{Visualization of adversarial loss landscapes. (a) shows the loss landscape of the baseline adversarial training method. (b) and (c) present the loss landscapes of our proposed method under different random directions. A visibly flatter loss profile can be observed in our method, indicating improved robustness against adversarial perturbations.}

    \label{fig:loss_landscape}
\end{figure}

Previous studies~\cite{jia2024improving}~\cite{shafahi2019adversarial} utilizing visualization techniques have demonstrated that a flatter adversarial loss landscape is positively correlated with improved model robustness against adversarial attacks. Therefore, we analyze and compare the adversarial loss landscapes of our proposed method with those of existing approaches~\cite{cui2021learnable}. Specifically, to generate the loss landscape, we follow the standard practice by selecting a random Rademacher direction (denoted as the $v$ direction) and an adversarial direction (denoted as the $u$ direction), which is obtained via the PGD-20 attack, as illustrated in Fig.~\ref{fig:loss_landscape}. In contrast to conventional adversarial training methods that employ only a single mean squared error (MSE) loss function, our method incorporates both a non-backpropagated MSE term and a root mean squared error (RMSE) component, resulting in a visibly flatter and more stable adversarial loss landscape.

% From the results, we observe that as \(\lambda\) increases from 1 to 2.5, the model achieves consistent improvements in robustness across all attack scenarios. 

% Specifically, the performance under PGD-10, PGD-20, and PGD-50 attacks improves significantly, and the robustness against stronger attacks such as C\&W and AA also shows noticeable enhancement. Therefore, based on the performance of these parameters, the peak performance is observed at \(\lambda = 2.5\), where the model achieves a clean accuracy of 86.05\% and demonstrates the highest robustness across all metrics. However, further increasing \(\lambda\) to 3 leads to a decline in both clean accuracy (85.18\%) and robustness. This suggests that while increasing \(\lambda\) can enhance robustness to a certain extent, an overly large value may compromise the balance between clean accuracy and adversarial robustness.  

\begin{table}[H]
\centering
\caption{Robustness (\%) on the CIFAR10 dataset using WRN34-10, with WideResNet as the guide model for both LBGAT and our AdaGAT-MSE methods. The best performance values are highlighted in bold.}
\label{table:ablation}
%\vspace{-3mm}
\begin{tabular}{lcccccc}
\toprule 
\setlength{\tabcolsep}{15pt}
$\lambda$  & PGD-10  & PGD-20  & PGD-50  & C\&W   & AA \\ 
\midrule
1   & 56.83 & 55.15 & 54.72 & 54.67 & 52.50 \\ 
2   & 57.45 & 56.40 & 55.98 & 55.44 & \textbf{53.71} \\ 
2.5 & \textbf{57.63} & \textbf{56.52} & \textbf{56.11} & \textbf{55.52} & 53.59 \\ 
3   & 56.97 & 55.97 & 55.66 & 54.40 & 53.49 \\

% $\lambda$    & Clean    & PGD-10  & PGD-20  & PGD-50  & C\&W   & AA \\ 
% \midrule
% 1  & 88.18 & 56.83 & 55.15 & 54.72 & 54.67 & 52.50 \\ 
% 2  & 87.11 & 57.45 & 56.40 & 55.98 & 55.44 & \textbf{53.71} \\ 
% 2.5 & \textbf{86.05} & \textbf{57.63} & \textbf{56.52} & \textbf{56.11} & \textbf{55.52} & 53.59 \\ 
% 3  & 85.18 & 56.97 & 55.97 & 55.66 & 54.40 & 53.49 \\
\bottomrule
\end{tabular}
%\vspace{0mm}
\end{table}

\section{Conclusion}
We propose an adaptive guidance strategy to enhance model robustness. Specifically, we introduce an additional loss function to constrain the guide model. During the training phase, the guide model and the target network are optimized synchronously, engaging in a dynamic interplay. The guide model generates real-time outputs to support the adversarial training of the target model. Consequently, the outputs produced by the guide model directly influence the robustness of the target model. The proposed loss function imposes constraints on the guide model during training, effectively improving the robustness of the target model.

Compared to previously proposed learnable methods, our approach demonstrates a stable increase in the robustness of the target model throughout training and achieves superior performance in later stages. Extensive experimental results validate the superiority of our method, showcasing its effectiveness in improving the model's robustness.
% \section*{Acknowledgment}

% The preferred spelling of the word ``acknowledgment'' in America is without 
% an ``e'' after the ``g''. Avoid the stilted expression ``one of us (R. B. 
% G.) thanks $\ldots$''. Instead, try ``R. B. G. thanks$\ldots$''. Put sponsor 
% acknowledgments in the unnumbered footnote on the first page.

% \section*{References}
\section*{Acknowledgment}
We acknowledge the support of the EPSRC-funded project National Edge AI Hub for Real Data: Edge Intelligence for Cyber-disturbances and Data Quality (EP/Y028813/1).

\bibliographystyle{splncs04}
\bibliography{cas-refs}
% \color{red}
% IEEE conference templates contain guidance text for composing and formatting conference papers. Please ensure that all template text is removed from your conference paper prior to submission to the conference. Failure to remove the template text from your paper may result in your paper not being published.

\end{document}